\documentclass[runningheads]{llncs}

\usepackage{stmaryrd}
\usepackage{listings}
\usepackage{lstautogobble}
\usepackage{capt-of}
\usepackage[T1]{fontenc}
\usepackage{textcomp}
\usepackage{amsmath}
\usepackage{xspace}
\usepackage{amsfonts}
\usepackage{amssymb}
\usepackage{enumerate}

\newcommand{\vx}[0]{\ensuremath{x}}
\newcommand{\vy}[0]{\ensuremath{y}}
\newcommand{\vz}[0]{\ensuremath{z}}
\newcommand{\vshape}[0]{\texttt{s}}
\newcommand{\vshapet}[0]{\ensuremath{t}}

\newcommand{\sh}[1]{\ensuremath{\texttt{sh}{:}{\texttt{#1}}}}

\newcommand{\conc}[0]{\ensuremath{\texttt{c}}}
\newcommand{\tripsquare}[3]{\ensuremath{{<}#1,\allowbreak #2,\allowbreak #3{>}}}

\newcommand{\SCL}[0]{\ensuremath{\texttt{\textbf{SCL}}}}
\newcommand{\gramdef}[0]{\ensuremath{:=}}

\newcommand{\hasshapenosubscript}[2]{\ensuremath{\Sigma(#1)}\xspace}
\newcommand{\hasshape}[2]{\ensuremath{\Sigma_{#2}(#1)}\xspace}

\newcommand{\sconst}{\texttt{s}}

\newcommand{\sigmagrammar}{\ensuremath{\varsigma}}
\newcommand{\isA}[0]{\texttt{isA}\xspace}

\newcommand{\hasshapePredicate}[0]{\ensuremath{\Sigma}\xspace}
\newcommand{\isIRI}[0]{\text{IRI}}

\newcommand{\tshacl}[0]{\textsc{shacl}\xspace}

\newcommand{\trdf}[0]{\textsc{rdf}\xspace}

\newcommand{\tiri}[0]{\textsc{iri}\xspace}

\usepackage{graphicx}

\usepackage{times}
\usepackage{soul}
\usepackage{url}
\usepackage{hyperref}

\usepackage[utf8]{inputenc}
\usepackage[small]{caption}
\usepackage{graphicx}

\usepackage{amsmath}

\usepackage{booktabs}
\usepackage{algorithm}
\usepackage{arevmath}
\usepackage[noend]{algpseudocode}
\usepackage{listings,multicol}
\usepackage{multicol}
\usepackage{longtable}
\usepackage{oz}
\usepackage{listings}
\usepackage{amsfonts}
\usepackage{relsize}
\usepackage{urwchancal}
\usepackage{stmaryrd}
\usepackage{ dsfont }
\usepackage{placeins}
\usepackage{longtable}
\usepackage{standalone}
\usepackage[htt]{hyphenat}
\usepackage{float}
\usepackage{mathtools}

\usepackage{xcolor}
\usepackage{listings}
\usepackage{lipsum}

\usepackage{listings}
\usepackage{lstautogobble}
\usepackage{color}
\usepackage{zi4}

\definecolor{bluekeywords}{rgb}{0.13, 0.13, 1}
\definecolor{greencomments}{rgb}{0, 0.5, 0}
\definecolor{redstrings}{rgb}{0.9, 0, 0}
\definecolor{graynumbers}{rgb}{0.5, 0.5, 0.5}

\usepackage[utf8]{inputenc}
\usepackage[T1]{fontenc}
\usepackage{zlmtt}
\usepackage{fancyvrb}
\usepackage{caption}

\usepackage{listings}
\newcommand{\shaclfol}[0]{SHACL2FOL}
\newcommand{\shacl}[0]{SHACL}

\newcommand{\fol}[0]{FOL}

\usepackage{amssymb}
\usepackage{xspace}
\usepackage{paralist}

\usepackage{amsthm}

\theoremstyle{remark}

\usepackage[final]{microtype}

\makeatletter
\def\orcidID#1{\smash{\href{http://orcid.org/#1}{\protect\raisebox{-1.25pt}{
\protect\includegraphics{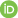}}}}}
\makeatother

\begin{document}

\title{\shaclfol: An FOL Toolkit for \\ SHACL Decision Problems}

\author{Paolo Pareti\orcidID{0000-0002-2502-0011} }

\authorrunning{P. Pareti}
\institute{University of Winchester, Winchester, United Kingdom \\ \email{paolo.pareti@winchester.ac.uk} }

\maketitle

\begin{abstract}
Recent studies on the \emph{Shapes Constraint Language} (\shacl), a
W3C specification for validating \trdf\ graphs, rely on translating the language
into first-order logic in order to provide formally-grounded
solutions to the \emph{validation}, \emph{containment} and \emph{satisfiability}
decision problems.
Continuing on this line of research, we introduce \shaclfol, the first automatic
tool that
\begin{inparaenum}[(i)]
  \item
    translates \shacl\ documents into \fol\ sentences and
  \item
    computes the answer to the two static analysis problems of satisfiability
    and containment; it also allow to test the validity of a graph with respect
    to a set of constraints.
\end{inparaenum}
By integrating with existing theorem provers, such as E and Vampire, the tool
computes the answer to the aforementioned decision problems and outputs the
corresponding first-order logic theories in the standard TPTP format.
We believe this tool can contribute to further theoretical studies of \shacl, by
providing an automatic first-order logic interpretation of its semantics, while
also benefiting \shacl\ practitioners, by supplying static analysis capabilities
to help the creation and management of \shacl\ constraints.

\end{abstract}

\section{Introduction}

Graphs are inherently schema-less data structures. However, practical applications of graph datasets typically require them to adhere to specific properties and constraints. This is crucial for data exchange, interoperability, and identifying data quality issues such as missing or corrupted information. The Shapes Constraint Language (\shacl) \cite{2017SHACL} is the W3C recommendation language for the validation of \trdf\ graphs. \shacl\ validation is based on a set of \emph{shapes} known as a \emph{shape graph}. Each shape defines specific constraints and identifies which nodes in a graph should be validated against these constraints. The graphs validated by \shacl\ are called \emph{data graphs}. A data graph (or simply graph) \emph{conforms} to a shape graph if it meets the constraints of all its shapes.

Representing \shacl\ shapes as a shape graph offers several practical advantages, but this format is not ideal for studying the semantics of these constraints. Consequently, recent theoretical studies on \shacl\ semantics have focused on translating shapes into formal languages \cite{SHACLstableModelSemantics,martin2020shapecontainment,pareti2022}. The first complete translation of \shacl\ that enabled the study of the decision problems of \emph{satisfiability} and \emph{containment} was achieved by translating it into a fragment of first-order logic called \SCL\ \cite{pareti2022}. Informally, a shape graph is \emph{satisfiable} if there exists a graph that conforms to it. A shape graph is \emph{contained} in another shape graph if every graph that conforms to the first also conforms to the second. 

Satisfiability and containment have several practical applications. For example, they can be used to detect errors in a shape graph, as an unsatisfiable set of shapes likely indicates a mistake. The containment decision problem is useful in optimization and management of shape graphs. Containment can also determine whether two shape graphs are equivalent by testing if each is contained in the other. 
The study in \cite{pareti2020} demonstrated that while satisfiability and containment decision problems are undecidable for the full language, they are decidable for several \shacl\ fragments. 
This work was extended in \cite{pareti2022} to account for \emph{recursive} constraints, a more complex type of constraints for which multiple semantics have been proposed \cite{Corman2018SHACL}.
These findings align with results from a separate group of researchers \cite{martin2020shapecontainment}, showing that the containment problem is decidable for a \shacl\ fragment representable using Description Logic. While previous work on Description Logic translations for \shacl\ mainly focuses on constraints on the edges (properties) of the data graph, the first-order logic approach used in this paper, namely \SCL , is the first to accurately model all of the \shacl\ core constraints components for static analysis tasks. Details of the differences between \SCL\ and other languages can be found in \cite{pareti2022b}.

Despite several articles demonstrating the decidability of these decision problems, an implementation was still missing until now. This paper addresses this gap by presenting \shaclfol , the first tool capable of solving the satisfiability and containment decision problems for \shacl .  
Additionally, \shaclfol\ is the first tool that translates both shape and data graphs into equivalent first-order logic theories for validation, satisfiability, and containment. We believe this translation can facilitate further research into both theoretical and practical applications of \shacl . Our tool achieves this by connecting \shacl\ with the extensive body of literature on first-order logic and providing a translation that enables practical experimentation. 
It should be noted that our tool can also compute the validity of a graph with respect to a shape graph. While our approach does not aim to outperform existing validators, it is significant because it demonstrates how even \shacl\ validation can be represented and studied as a first-order logic satisfiability problem. 

\section{Background}

In \shacl\ a shape graph consists of a set of shapes. These shapes impose restrictions on the structure of a graph by specifying constraints that must be met by certain nodes, known as \emph{target nodes}. A shape thus consists of 1) its name, that acts as a unique identifier, 2) a target definition, comprising a set of target declarations, each represented by a unary query identifying \trdf target nodes that must satisfy its constraints; (3) a set of constraints. The \shacl\ specification includes various constraint types, known as \emph{constraint components}. For example, the \sh{datatype} component restricts an \trdf term to be an \trdf literal of a specific datatype. A graph conforms to a shape graph if all constraints of all shapes in the shape graph are satisfied by the target nodes of the corresponding shapes.

We will now briefly outline the syntax of the \SCL\ language, and how data graphs are represented in first-order logic.  For a more comprehensive explanation, please refer to \cite{pareti2022}. In  \SCL , triples of an \trdf\ graph are modelled as binary relations, with atom $R(s,o)$ corresponding to triple $\tripsquare{s}{R}{o}$. The \emph{inverse} role is identified by a minus sign, so  $R^{-}(s,o) = R(o,s)$. The binary relation name \isA represents class membership triples $\tripsquare{s}{\texttt{rdf:type}}{o}$ as $\isA(s,o)$. Sentences in the \SCL\ language follow the $\varphi$ grammar outlined in Definition \ref{def:synGeneral}. 

\begin{table}[t]
\begin{center}
\caption{ 
Translation of a shape with name $\vshape$ with a target definition $\vshapet$, into an \SCL\ target axiom.
} 
\label{tab:translationTargs}
\begin{tabular}{ |l | l |}
 \hline
 Target declaration in $\vshapet$  & \SCL\ target axiom \\ \hline 
 Node target (node \conc) & $\hasshape{\conc}{\vshape}$ \\ \hline
 Class target (class \texttt{c})  & $\forall \vx . \isA(\vx,$\texttt{c}$)\rightarrow \hasshape{\vx}{\vshape}$\\ \hline
 Subjects-of target (relation $R$)  & $\forall \vx, \vy . R(\vx,\vy) \rightarrow \hasshape{\vx}{\vshape}$\\ \hline
 Objects-of target (relation $R$)  & $\forall \vx, \vy . R^{-}(\vx,\vy) \rightarrow \hasshape{\vx}{\vshape}$ \\ \hline
\end{tabular}
\end{center}
\end{table}

\begin{definition}
    \label{def:synGeneral}
    The \emph{SHACL} first order language (\SCL, for short) is the set of first order \emph{sentences} built according to the following context-free grammar, where $\conc$ is a constant from the domain of \trdf terms, $\hasshapePredicate$ is a shape relation, $F$ is a filter relation, with shape relations disjoint from filter relations, 
    $R$ is a binary-relation name, $^{\star}$ indicates the transitive closure of the relation induced by $\pi(\vx,\vy)$, the superscript $\pm$ refers to a relation or its inverse, and $n \in \mathbb{N}$.
    \begin{align*}
\varphi \gramdef\; & 
      \top
      \mid \varphi \wedge \varphi
      \\ & \mid \hasshapenosubscript{\conc}{\sconst} 
      \mid \forall \vx \,.\, \isA(\vx, \conc) \rightarrow \hasshapenosubscript{\vx}{\sconst} 
      \mid  \forall \vx, \vy \,.\, R^{\pm}(\vx, \vy) \rightarrow \hasshapenosubscript{\vx}{\sconst} 
      \\ & \mid \forall \vx . \; \hasshapenosubscript{\vx}{\sconst}  \leftrightarrow \psi(\vx) ; \\
\psi(\vx) \gramdef\;
        & \top 
        \mid \neg \psi(\vx)
        \mid \psi(\vx) \wedge \psi(\vx) 
        \mid \vx = \conc 
        \mid F(\vx)
        \mid \hasshapenosubscript{\vx}{\sconst}  
        \mid \exists \vy .\, \pi(\vx, \vy) \wedge \psi(\vy)
        \\ & \mid \neg \exists \vy .\, \pi(\vx, \vy) \wedge R(\vx, \vy)
          & 
        \\ & \mid \forall \vy .\, \pi(\vx, \vy)
          \leftrightarrow R(\vx, \vy)  & 
        \\ & \mid \forall \vy, \vz \,.\, \pi(\vx, \vy) \wedge R(\vx, \vz) \rightarrow \sigmagrammar(\vy, \vz) & 
        \\ & \mid \exists^{\geq n} \vy \,.\, \pi(\vx, \vy) \wedge \psi(\vy)  ; &  \\
\pi(\vx, \vy) \gramdef\;
        & R^{\pm}(\vx, \vy) 
        \\ & \mid \exists \vz \,.\, \pi(\vx, \vz) \!\wedge\!
          \pi(\vz, \vy) & 
        \\ & \mid \pi(\vx, \vy) \!\vee\! \pi(\vx, \vy) &  
        \\ & \mid  (\pi(\vx, \vy))^{\star}  &   
        \\ & \mid \vx \!=\! \vy \!\vee\! \pi(\vx, \vy) ; &  \\
\sigmagrammar(\vx, \vy) \gramdef\;
        & \vx <^{\pm} \vy \mid \vx \leq^{\pm} \vy.  
    \end{align*}
\end{definition}

Symbol $\varphi$ corresponds to a shape graph. An \SCL\ sentence could be empty ($\top$), a conjunction of shape graphs, a \emph{target axiom} representing a target definition (a production of the 3rd, 4th and 5th production rule) or a \emph{constraint axiom} representing a constraint (a production of the last production rule). Target axioms take one of three forms, based on the type of target declarations. The translation of \tshacl target declarations into \SCL\ target axioms is summarised in Table~\ref{tab:translationTargs}. The non terminal symbol $\psi(\vx)$ corresponds to the subgrammar of the \tshacl constraints components. Within this subgrammar, $\top$ identifies an empty constraint, $\vx = \conc$ a constant equivalence constraint and $F$ a monadic filter relation (e.g.\ $F^{\isIRI}(\vx)$, true iff $\vx$ is an IRI). Constraints on individual \tiri\ or literal nodes are captured by $F(\vx)$ and the $\sigmagrammar(\vx, \vy)$  subgrammar. The $\pi(\vx, \vy)$ subgrammar models \tshacl property paths. Within this subgrammar the production rules denote, respectively, sequence paths, alternate paths, zero-or-one path and zero-or-more path.

\section{The SHACL2FOL Toolkit}

This section introduces the \shaclfol\ software. The code is available at the link below.\footnote{\url{https://w3id.org/shacl2fol}, original Github repository: \url{https://github.com/paolo7/SHACL2FOL}} This software uses the TPTP language \cite{sutcliffe1998tptp}, and it was tested on compatible theorem provers such as E \cite{schulz2002brainiac} and Vampire \cite{riazanov2001vampire}. Our tool can be configured to perform one of three tasks: satisfiability checking, containment checking and validation. Instructions on how to run this tool are available in its documentation. 

\subsection{Satisfiability Checking}

The main functionality of \shaclfol\ involves translating a shape graph into a TPTP file. This translation follows the mapping $\tau$ from \shacl\ to \SCL\ defined in \cite{pareti2022}. Specifically, given a shape graph $A$, $\tau(A)$ is an \SCL\ sentence that is satisfiable if and only if $A$ is satisfiable. When checking satisfiability, \shaclfol\ takes a single shape graph as an input, and generates a TPTP file that serialises $\tau(A)$. Satisfiability is then verified using one of the aforementioned theorem provers.

It should be noted that the \SCL\ language uses the Unique Name Assumption (UNA), which our tool can encode using one of two approaches. In the first approach, UNA is encoded as the pairwise inequality of all known constants. However, this method results in an axiomatization that grows polynomially with the number of known constants. Therefore, it might not be suitable for large data graphs. 
The second approach makes use of the ``\$distinct'' type  of Typed First-Order Form with Arithmetic (TFF) \cite{tff}. Elements of this type are by definition unequal to each other. Since each constant only needs to be declared of this type once, this approach scales linearly with the number of known constants, making it more suitable for validating large graphs.
It should be noted that our translation does not use any TFF features (such arithmetic) that are not already part of standard first order logic, with the exception of the ``\$distinct'' type.

From a practical standpoint, it is worth noting that a shape graph might be satisfiable even if some of its shapes are not (i.e.\ if there cannot exist a node that satisfies that shape). It is possible, in fact, that the shape graph is satisfiable only on an empty graph. This can be considered a trivial case, which one might want to exclude when testing for satisfiability. One way to test for a stronger notion of satisfiability is to introduce additional constraints in the shape graph, targeting dummy nodes, to require them to conform to specific shapes. Essentially, this approach can be used to require each shape (or a subset thereof) to have at least one node conforming to it.

\subsection{Containment Checking}

Checking for containment follows a similar approach as above, but this time \shaclfol\ takes two shape graphs $A$ and $B$ as inputs, and outputs the serialisation of a first order sentence that is unsatisfiable if and only if $A$ is contained in $B$. The specific approach to reduce this containment check to a satisfiability one is detailed in Lemma 4 of \cite{pareti2022}. Informally, this sentence is a conjunction of $\tau(A)$ with a sentence derived from $\tau(B)$ by negating its target axioms, that is, the axioms that formalise the \shacl\ target declarations. Intuitively, this check looks for a data graph where the first shape graph is satisfied, while the second one is not.

\subsection{Validation Checking}

When performing validation, \shaclfol\ takes a shape graph $A$ and a data graph $G$ as inputs, and generates a serialisation of a first order sentence that is satisfiable if and only if the data graph conforms to the shape graph. This is achieved by creating a first-order logic sentence that is the conjunction of $\tau(A)$ and the serialisation $\tau^{+}(G) \wedge \tau^{-}(G)$ of the data graph. The purpose of this serialisation is to ensure that the interpretation of binary relations representing \trdf\ properties matches the triples of the graph. In other words, it ensures that in a first-order instance that satisfies the output sentence, for each relation name $R$ and pair of constants $s$ and $o$,  pair $(s,o)$ is in the interpretation of $R$ if and only if triple $\tripsquare{s}{R}{o} \in G$.

Let $\hat{R}$ be the set of all relation names in $A$ and $G$. Sentences $\tau^{+}(G)$ and $\tau^{-}(G)$ are defined as follows.


\[
\tau^{+}(G) = \bigwedge_{\forall \tripsquare{x}{R}{y} \in G} R(x,y)
\]
\[
\tau^{-}(G) = \bigwedge_{\forall R \in \hat{R}} \begin{cases} 
\neg \exists x, y . \; R(x,y) \text{ if } \neg \exists x, y . \; \tripsquare{x}{R}{y} \in G  & \\
\forall x, y . \; R(x,y) \rightarrow ( \bigvee_{\forall z, k . \; \tripsquare{z}{R}{k} \in G} x = z \wedge y = k ) \text{ otherwise }
\end{cases}
\]

Intuitively, $\tau^{+}(G)$  ensures that a valid interpretation models all the triples in the graph, and $\tau^{-}(G)$ ensures that it does not model any triples not in the graph.

\subsection{Node Constraints}

The \SCL\ language captures all of the \shacl\ core constraint components, including the constraints on individual nodes such as those on the length of string literals, on node types and on datatypes. A correct implementation of a satisfiability and containment checker for \shacl\ must account for their non-trivial interactions. For instance, there exists only a finite number of constants that are of datatype integer and fall within the range of, say, greater than 2 and less than 5. This finite number might determine whether a \shacl\ cardinality constraint is satisfiable or not.

An implementation of the more complex \shacl\ constraints on literals in is possible but time consuming. For this reason, among the constraints on literals, only the \sh{NodeKind}, constraint has been implemented. The \sh{in} and \sh{hasValue} are also trivially supported as they can be modelled with the constant equality operator.

\subsection{Testing}

Preliminary development testing has been successfully performed on small shape and data graphs. Due to the absence of an established benchmark of \shacl\ shapes, comprehensive scalability testing remains a non-trivial task and it is left for future work. Nevertheless, the initial results indicate that the tool can effectively handle a small number of shapes, thus making it suitable for sub-problems such as constraint satisfiability and constraint containment \cite{pareti2022}.

\section{Conclusion}

This paper has introduced the \shaclfol\ tool, which represents the first practical implementation of a satisfiability and containment checker for \shacl\ based on \SCL . The \SCL\ language is the only formalization of SHACL capable of expressing all \shacl\ core constraint components for the aforementioned static analysis tasks. The key details of this tool have been discussed, and the source code along with documentation is available in an open-source repository.
Future work will focus on completing its implementation and integrating it with existing \shacl\ validators and management tools. One of the aims of this research is to bridge recent theoretical results into practical tools that can assist \shacl\  practitioners. We also hope that this implementation will enable further research into additional static analysis problems, such as the interactions of \shacl\ with ontologies and inference rules \cite{pareti2019c}, or with database transformations \cite{boneva2023static}.

\bibliographystyle{splncs04}
\bibliography{litbib}

\end{document}